%% file: acl2020.tex
\documentclass[11pt,a4paper]{article}
\usepackage[dvipsnames]{xcolor}
\usepackage[hyperref]{acl2020}
\usepackage{times}
\usepackage{url}
\usepackage{latexsym}
\usepackage{mathtools}
\usepackage{bbm}
\usepackage{dutchcal}
\usepackage{graphicx}
\usepackage{amsmath}
\usepackage{subcaption}

\usepackage{microtype}
\usepackage{multirow}
\usepackage{pifont}
\usepackage{lineno}
\usepackage[normalem]{ulem}
\usepackage[bottom]{footmisc}
\usepackage{float}

\newcommand{\dataset}[1]{\textsc{#1}\xspace}

\usepackage{microtype}
\input{math_comm.tex}

\aclfinalcopy % Uncomment this line for the final submission
\def\aclpaperid{16} %  Enter the acl Paper ID here

\title{Domain Adaptative Causality Encoder}

\author{Farhad Moghimifar$^1$, \bf{Gholamreza Haffari}$^2$ \and \bf{Mahsa Baktashmotlagh}$^1$ \\
         $^1$The School of ITEE, The University of Queensland, Australia\\
         $^2$Faculty of Information Technology, Monash University, Australia\\
         \tt \{f.moghimifar,m.baktashmotlagh\}@uq.edu.au\\
         \tt gholamreza.haffari@monash.edu, 
         }

\date{}

\begin{document}
\maketitle

\begin{abstract}
    \input{00_abstract.tex}
\end{abstract}

\section{Introduction}
\label{sec:intro}
\input{01_introduction.tex}

\section{Related Works}
\label{sec:related_works}
\input{02_related_works.tex}

\section{Our Approach}
\label{sec:models}
\input{03_models.tex}

\section{Experiments}
\label{sec:experiment}
\input{04_experiments.tex}

\section{Conclusion}
\label{conclusion}
\input{05_conclusion.tex}

\clearpage
\newpage

\bibliography{ref.bib}
\bibliographystyle{acl_natbib}

\end{document}

%% file: math_comm.tex
\usepackage{amsthm,amsmath,amsfonts,bm,xspace}
\usepackage{upgreek}
\usepackage{color}

\def\vs{{\em v.s.}\xspace}
\newcommand{\comment}[1]{}

\def\eqref#1{(\ref{#1})}
\def\1{\bm{1}}

\def\vA{{\bm{A}}}
\def\vX{{\bm{X}}}
\def\vY{{\bm{Y}}}

\def\vtheta{{\bm{\theta}}}

\def\vb{{\bm{b}}}

\def\vf{{\bm{f}}}

\def\vh{{\bm{h}}}

\def\vs{{\bm{s}}}

\def\vx{{\bm{x}}}
\def\vy{{\bm{y}}}

\DeclareMathAlphabet{\mathsfit}{\encodingdefault}{\sfdefault}{m}{sl}
\SetMathAlphabet{\mathsfit}{bold}{\encodingdefault}{\sfdefault}{bx}{n}
\newcommand{\tens}[1]{\bm{\mathsfit{#1}}}

\def\tW{{\tens{W}}}

%% file: 00_abstract.tex
%Causal relationships form the basis for reasoning and decision-making in Artificial Intelligence systems. To exploit the large volume of textual data, 
Automated discovery of causal relationships from text is a challenging task.
Current approaches which are mainly based on the extraction of low-level relations among individual events are limited by the shortage of publicly available labelled data. Therefore, the resulting models perform poorly when applied to  a  distributionally different domain for which labelled data did not exist at the time of training.
%
%limited by the shortage of publicly available labeled data needed to extract low-level relations among individual events for this task. Furthermore, they 
%
%Current approaches  are limited to the extraction of low-level relations among individual events on a existing specific dataset, and perform poorly when used for causal identification in a  new domain. % for causal iin  case of domain shift. 
%To overcome the limitations of the existing approaches, 
To overcome this limitation, in this paper, we leverage the characteristics of dependency trees and adversarial learning to address the tasks of adaptive causality identification and localisation. The term adaptive is used since the training and test data come from two distributionally different datasets, which to the best of our knowledge, this work is the first to address. Moreover, we present a new causality dataset, namely \dataset{MedCaus}\footnote{https://github.com/farhadmfar/ace}, which integrates all types of causality in the text.
Our experiments on four different benchmark causality datasets demonstrate the superiority of our approach over the existing baselines, by up to 7\% improvement, on the tasks of identification and localisation of the causal relations from the text.

%% file: 01_introduction.tex
Causality is the basis for reasoning and decision making. While human-beings use this psychological tool to choreograph their environment into a mental model to act accordingly \cite{pearl2018book}, the inability to identify causal relationships is one of the drawbacks of current Artificial Intelligence systems \cite{lake2015human}. The projection of causal relations in natural language enables machines to develop a better understanding of the surrounding context and helps downstream tasks such as question answering~\cite{hassanzadeh2019answering},  text summarisation~\cite{ning2018joint}, and natural language inference~\cite{roemmele2011choice}.

The task of textual causality extraction can be divided into two main subtasks, causality identification and causality localisation. The former subtask focuses on identifying whether a sentence carries any causal information or not, which can be seen as classification problem. The objective of the latter subtask is to extract text spans related to cause and effect, subject to existence.

The automatic identification and localisation of causal relations in textual data is considered a non-trivial task \cite{dasgupta2018automatic}. Causal relations in text can be categorised as marked/unmarked and explicit/implicit \cite{blanco2008causal,hendrickx2009semeval}. Marked causality refers to the case where a causal linguistic feature, such as ``because of'', is stated in the sentence. For example, in \textit{``His OCD is because of genetic factors."}, \textit{because of } is a causal marker,
whereas in unmarked causality there is no such indicator. 
For instance, in \textit{``Don't take these medications before driving. you might feel sleepy."} the cause and effect relationship is spread between two sentences without a marker.
On the other hand, explicit causality refers to the case where both cause and effect are mentioned in text. However, in implicit causality, either cause or effect are directly mentioned in the text. In a more complex case called nested causality, multiple causal relations may exist in one sentence (e.g.,\ \textit{``Procaine can also cause allergic reactions causing individuals to have problems with breathing"}). All of these ambiguities contribute to the challenging nature of this task.% Therefore, identifying causality is a challenging task ~\citep{dasgupta2018automatic}.
%The causal relations between events or concepts can be stated in explicit or implicit ways. For instance, in `` He had knee surgery because of the rupture in his anterior cruciate ligament", the connective phrase \emph{because of} explicitly indicates causal relation. Whereas, in ``After taking this medicine, you shouldn't drive. You might feel sleepy." there is an implicit causal relation between two sentences. Also, in some cases, the usual causal connective phrases may imply temporal relation rather than causal. Hence, identifying causality is a challenging task. 

Traditional approaches to address the problem of causality extraction mainly relied on predefined linguistic patterns and rules to identify the existence of causal relations in a sentence \cite{mirza2016catena}. More advanced approaches combined pattern-based methods with machine learning techniques \cite{zhao2018causaltriad}, and as such they require heavy manual feature engineering to perform reasonably. To overcome this problem, the recent approaches have adopted deep learning techniques to extract meaningful features from the text~\cite{liang2019multi,martinez2017neural}. 

However, all the aformentioned approaches suffer from the problem of domain shift, where there is a distribution difference between the training and the test data. More specifically, the existing approaches perform poorly on the data from a new test domain (e.g. financial) which is contextually different from the training domain (e.g. medical). 
%\reza{where they do not well on data from a different domain with distributional mismatch to the source domain?}
%where the domains of data distributionally mismatch. 

To overcome the limitations of existing approaches on the tasks of causality identification and localisation, we propose a novel approach for domain-adaptive causality encoding which performs equally well when applied on the out-of-domain sentences. Our contribution is three-fold:

\begin{itemize}
    \item To identify causal relationships and extract the corresponding causality information within a sentence, using graph convolutional networks, we propose a model which takes into account both syntactic and semantic dependency of words in a sentence. Extensive experimental results suggest that our proposed models for causality identification and localisation outperform the state-of-the-art results.
    \item We propose to use a gradient reversal approach to minimise the distribution shift between the training and test datasets. Our proposed adaptive approach improves the performance of the existing baselines by up to 7\% on the tasks of adaptive causality identification.
    \item To fill the gap of the current causality datasets on encompassing different types of causality, we introduce \dataset{MedCaus}, a dataset of 15,000 labelled sentences, retrieved from medical articles. This dataset consists of sentences 
%with simple and complex syntactic structure, 
with labels of explicit, implicit, nested, and no causality. 

\end{itemize} 
%We evaluate our  methods on our proposed dataset as well as the public datasets, and observe that in comparison to the state-of-the-art models, our systems achieve better results. \reza{be more specific about the numbers}

%To recap, our contribution in this work is three-fold: 1) we present \dataset{MedCaus}, a labelled dataset consisting of 15,000 sentences retrieved from medical articles. 2) We propose a Graph Convolutional based network for identifying causality in sentences. and 3) We propose an domain adaptive method  to enable causal identification for out-of-domain sentences. We tested our proposed models on our proposed dataset as well as the public datasets and observed that in comparison to the state-of-the-art models, our systems achieve better results.

%% file: 02_related_works.tex
%Causal relations in text can be categorised as marked/explicit or unmarked/implicit \cite{blanco2008causal,hendrickx2009semeval}. Marked causality refers to the case where a causal linguistic feature is stated in the sentence. For example, in \textit{``His OCD is because of genetic factors."}, \textit{because of } is a causal marker, whereas in unmarked causality there is no such indicator. For instance, in \textit{``Don't take these medications before driving. you might feel sleepy."} the cause and effect relationship is spread between two sentences without a marker. 

%On the other hand, explicit causality refers to the case where both cause and effect are mentioned in text. Whereas, in implicit causality, either cause or effect are not directly mentioned in the text. 
%In a more complex case, multiple causal relations may exist in one sentence (e.g.,\ ``Procaine can also cause allergic reactions causing individuals to have problems with breathing"). Therefore, identifying causal relations from text can be very challenging.

The projection of causal relation in textual data can be in various forms, depending on the type of causality. The categorisation mentioned in Section \ref{sec:intro} can indicate the relation between pairs of events, phrases, concepts, named entities or a mixture of the aforementioned text spans~\citep{hashimoto2019weakly}. Some works in the area have endeavoured to extract and present the textual information between concepts or events.
Causal relations are a component of SemEval task \citep{hendrickx2009semeval}, but it involves a limited set of causal relations between pairs of nominals. \citet{do2011minimally} developed a framework based on combining semantic association and supervised causal discourse classification in order to identify causal relations between pairs of events. They expand the patterns in \emph{pdtb}~\citep{lin2009recognizing} using a self-training approach. 
Other methods~\citep{riaz2014recognizing,riaz2013toward} leveraged linguistic features such as part-of-speech information, alongside with discourse markers, for identifying causal relations between events. \citet{an2019extracting} used the syntactic patterns and word vectors to develop an unsupervised method for constructing causal graphs. To expand the repository of causal syntactic patterns, \citet{hidey2016identifying} built a parallel corpus between English Wikipedia and Simple Wikipedia, where the same causal relation might be in different syntactic markers in two parallel sentences. A supervised method was adapted by \citet{mirza2016catena} using lexical, semantic, and syntactic features within a sentence to address this task.

Using \citet{hidey2016identifying}'s method, \citet{martinez2017neural} created a set of labelled sentences, assuming all of the sentences include a causal relation, and presented a neural model based on LSTM to identify causality. \citet{dasgupta2018automatic} collected a dataset and developed a model using BiLSTM for extracting causal relation within a sentence\footnote{The source code and dataset are not publicly available.}. Other approaches in event prediction applied Granger causality \citep{granger1988some} to identify causal relations in time series of events \citep{kang2017detecting}. \citet{rojas2017causal} defined a proxy variable, which may carry some information about cause and effect, to identify causal relationship between static entities. \citet{zhao2017constructing} developed a causality network embedding for event prediction. \citet{de2017causal} proposed a convolutional neural network model for identifying causality.
\citet{liang2019multi} also deployed a self-attentive neural model to address the task of causality identification, however, the extraction of causal information is not addressed by their model. 

In more recent works, a dataset of counterfactual sentences was released, as a part of SemEval2020 Task5 \citep{yang2020semeval}. The aim of this task is to identify and tag the existing counterfactual part of the sentence. Some works have attended to address this task using different deep learning architectures \citep{patil2020cnrl,abi2020semeval}. While counterfactuals are usually represented in form of a causal relation, this dataset does not cover different forms of textual causality. 

As opposed to the aforementioned models, we propose a unified neural model for addressing both tasks of identifying and localising causality from a sentence. Our method leverages both syntactic and semantic relations within a sentence, and adapts to out-of-domain sentences.

%% file: 03_models.tex
\begin{figure*}[t]
    \vspace{-2ex}
    \centering
    \includegraphics[width=\linewidth, height = 0.24\linewidth]{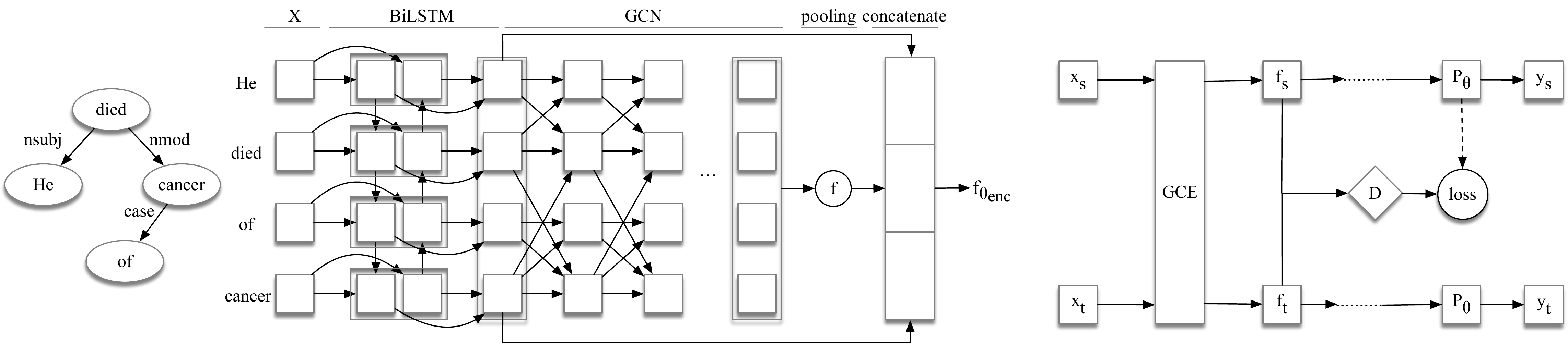}
    \caption{(Left) The dependency tree of the sentence \textit{``He died of cancer.''}.
    %obtained from   Stanford CoreNLP \citep{manning-EtAl:2014:P14-5}. 
    (Middle) Architecture of our proposed models for Causality Identifier~(GCE), which uses the retrieved dependency tree as its core. (Right) The architecture of our proposed Adaptive Causality Encoder~(ACE), which uses the structure of GCE as feature extractor.}
    \label{fig:arch}
\end{figure*}

In this section, we first describe the architecture of causality extractor, which uses graph convolutional networks (GCN) at its core. We then present how we make use of the adversarial learning strategy for adapting the model to new domains. Figure \ref{fig:arch} illustrates the high level overview of our approach.

\subsection{Graphical Causality Encoder (GCE)}
Given a sentence $\vX = [\vx_1,\dots,\vx_n]$, where $\vx_i$ is the vector representation of the $i$-th token of the sentence, the goal of our model is two-fold: identifying whether or not the causal relation exists, and locating the position of cause and effect in the given sentence.

The core part of our causal identification model consists of an \emph{L}-layer graph convolutional  network~(GCN) which takes as input the dependency tree of a sentence,  obtained through Stanford CoreNLP~\cite{manning2014stanford}. The dependency tree can be represented with an $n \times n$ adjacency matrix $\vA$, where \emph{n} is the number of nodes in the graph. In the adjacency matrix, $A_{ji} = A_{ij} = 1$  if an edge connects $i$ to $j$, and zero otherwise. 
Given $\vh_i^{(l-1)}$  as the representation of the node $i$ at layer $l-1$, GCN updates the node representation at layer $l$ as follows ~\citep{zhang2018graph,kipf2016semi}:
%\vspace{-2ex}
\begin{equation}
    \small
    \vh_i^{(l)} = \vf^{\textrm{actv}}(\sum_{j=1}^n \tilde{A}_{ij}\tW^{(l)}\vh_j^{(l-1)}/d_i + \vb^{(l)})
    \label{eq_gcn}
\end{equation}
where $\tilde{\vA} = \vA + \bm{I}$, with $\bm{I}$ the identity matrix, $\vf^{\textrm{actv}}$ an activation function (i.e., element-wise RELU), $\vb^{(l)}$ the bias vector, $\tW^{(l)}$ the weight matrix, and $d_i = \sum_{j=1}^n\tilde{A}_{ij}$ the degree of node $i$.

This formation captures the hidden embeddings of each token in a sentence with respect to its neighbours with maximum distance of \emph{L}, with \emph{L} the number of GCN layers. To take the words order and disambiguity into account and make the model less prone to errors from the dependency relations' results, we feed the word vectors into a bi-directional long short-term memory (BiLSTM) network. The output of the BiLSTM $\vh^{(0)}_i$ is then used in the GCN, as illustrated in Equation \ref{eq_gcn}. Hence, after applying the BiLSTM and GCN, each sentence is represented as:
\begin{equation}
    \small
    \vh_{\textrm{GCN}}(\vX) =  \vf^{\textrm{pool}}(\textrm{GCN}(\textrm{BiLSTM}(\vX))
    \label{eq:sent_rep}
\end{equation}
where $\vf^{\textrm{pool}} : \mathbbm{R}^{d\times n} \to \mathbbm{R} ^d$ is a pooling function generating the  representations for the $n$ tokens of the sentence.
The final sentence representation is obtained by a feed forward network (FFNN) whose input is the concatenation of $\vh_{\textrm{GCN}}(\vX)$ and $\vh_{\textrm{BiLSTM}}(\vX)$. Note that $\vh_{\textrm{BiLSTM}}(\vX)$ is the contextualised representation of the sentence from BiLSTM which is constructed by concatenating the leftmost and rightmost hidden states:
\begin{equation}
    \small
    \vf_{\vtheta_{\textrm{enc}}}(\vX) = \textrm{FFNN}([\vh_{\textrm{GCN}}(\vX);\vh_{\textrm{BiLSTM}}(\vX)])
\end{equation}
where $\vtheta_{\textrm{enc}}$ contains the collection of  parameters of the GCN, BiLSTM, and the feed-forward network. 
This representation is then used to address the two main sub-tasks:
\begin{itemize}
    \item For Task1, which is identifying causal relation within a sentence,
    we use this representation to get the probability of output classes,
    \begin{equation}
        \small
        P_{\vtheta_{\textrm{class}}}(\textrm{causality}|\vX) = \sigma(\vf_{\vtheta_{\textrm{enc}}}(\vX) . \tW_{\textrm{class}} + b_{\textrm{class}})
    \end{equation}
where $\vtheta_{class}:=\{\tW_{\textrm{class}},b_{\textrm{class}}\}$ contains the classifier's parameters, and $\sigma$ is the sigmoid function. 
    \item For Task2, locating cause and effect in a sentence, we use this representation to obtain the probability of the corresponding tag for each token. Since there are strong dependencies across tags, by adopting conditional random fields, we model the tagging decision jointly, with respect to surrounding tags. Consider $\vY = [\vy_1,\dots,\vy_n]$ the sequence of tag predictions. The score corresponding to this sequence is defined as:\\
\begin{equation}
    \small
    \vs(\vX,\vY) = \sum_{i=0}^n\vf_{\vtheta_{\textrm{enc}}}(\vX)_{i,\vy_i} + \sum_{i=0}^nT_{\vy_i,\vy_{i+1}}
    \label{eq:score}
\end{equation}
where $T$ is a square matrix with its size corresponding to the number of distinct tags. $T_{i,j}$ is representative of the score related to shifting from tag \emph{i} to tag \emph{j}. The probability of the tag sequence $\vY$ given $\vX$ is then defined as~($\vY_\vX$ which denotes all possible sequences of tags for $\vX$):\\
\begin{equation}
    \small
    P_{\vtheta_{\textrm{seq}}}(\vY|\vX) = \vs(\vX,\vY) - log(\sum_{\vy \in \vY_{\vX}} e^{s(\vX,\vy)})
\end{equation}
Here, ${\vtheta_{\textrm{seq}}}$ contains the sequence tagger's parameters.
\end{itemize}
%We then use this representation to get the probability of output classes, 
%$$P_{\vtheta_{\textrm{class}}}(\textrm{causality}|\vX) = \sigma(\vf_{\vtheta_{\textrm{enc}}}(\vX) . \tW_{\textrm{class}} + b_{\textrm{class}})$$
%where $\vtheta_{class}:=\{\tW_{\textrm{class}},b_{\textrm{class}}\}$ contains the classifier's parameters, and $\sigma$ is the sigmoid function. 

\subsection{Adaptive Causality Encoder (ACE)}
In this section, we represent a domain adversarial approach to adaptive causality identification and localisation. In unsupervised domain adaptation, we are given a source labelled data $\mathnormal{D}_s = \{(\vX^{(i)}_s, Y^{(i)}_s)\}_{i=1}^{n_s}$ and unlabelled target data $\mathnormal{D}_t = \{\vX^{(j)}_t\}_{j=1}^{n_t}$. Our aim is to reduce the distributional shift between the two domains, and predict the labels of the target domain. 
Inspired by \citep{ganin2016domain,long2018conditional}, we make use of an adversarial learning strategy, where the domain discriminator is trained to distinguish the source and domains, while the feature representation is trained to confuse the domain discriminator.

More formally, let us consider the following domain classifier,
\begin{equation}
    \small
    P_{\vtheta_{\textrm{dom}}}(\textrm{source}|\vX) = \sigma(\vf_{\vtheta_{\textrm{enc}}}(\vX) . \tW_{\textrm{dom}} + b_{\textrm{dom}})
\end{equation}
where $\vtheta_{dom}:=\{\tW_{\textrm{dom}},b_{\textrm{dom}}\}$ is the domain classifier's parameters.
Our domain adversarial training objective is defined as,

\vspace{-5mm}
{\small
\begin{align}
\mathcal{L}(\vtheta_{\textrm{enc}},\vtheta_{\textrm{dom}},\vtheta_{\textrm{task}}) := \sum_{(\vX,Y) \in D_s} \log P_{\vtheta_{\textrm{class}}}(Y|\vf_{\vtheta_{\textrm{enc}}}(\vX)) \nonumber \\
 - \sum_{\vX \in D_s} \log P_{\vtheta_{\textrm{dom}}}(\textrm{source}|\vf_{\vtheta_{\textrm{enc}}}(\vX)) \nonumber \\
 - \sum_{\vX \in D_t} \log \big( 1 - P_{\vtheta_{\textrm{dom}}}(\textrm{source}|\vf_{\vtheta_{\textrm{enc}}}(\vX)) \big). \nonumber
\end{align}
}
The model parameters are then trained by,
$$\arg\max_{\vtheta_{\textrm{task}}}
\max_{\vtheta_{\textrm{enc}}}
\min_{\vtheta_{\textrm{dom}}}
\mathcal{L}(\vtheta_{\textrm{enc}},\vtheta_{\textrm{dom}},\vtheta_{\textrm{task}})$$
where minimization over $\vtheta_{\textrm{dom}}$ strengthens the accuracy of the domain classifier, but maximizing over $\vtheta_{\textrm{enc}}$ tries to confuse the domain classifier and strengthen the causality classifier. 

\subsection{Tagging Scheme}

The objective of the task of causality localisation is to assign a label to each token in a sentence to locate the position of cause and effect. Cause and effect of a causal relation may span several tokens in a sentence. Therefore, the labels of a sentence usually are represented in the IOB-format (Inside, outside, and beginning). In this format, B-\emph{label} indicates beginning of the span \emph{label}, I-\emph{label} shows a token inside the \emph{label} but not the first token, and O-\emph{label} represents the token as an outsider of \emph{label}. However, inspired by \citep{ratinov2009design} and \citep{dai2015enhancing}, we use IOBES, an extended version of IOB, which also accounts for singleton labels and end of the label span token. Furthermore, to keep the tags consistent with the Equation \ref{eq:score}, we add a \emph{start} and \emph{end} label to the set of tags.

%The proposed architecture consists of three parts: a deep feature extractor $\mathnormal{G}_f$ (the $h_{sent}$ obtained from  \ref{eq:sent_rep}), a label predictor $\mathnormal{G_y}$ obtained from feeding the output of $\mathnormal{G}_f$ to a linear layer and softmax, and a domain classifier $\mathnormal{G}_d$. By confusing the domain classifier $\mathnormal{G}_d$ through a gradient reversal layer, we aim to minimize the distribution difference between the training and test data so that our model can perform similarly well on the test domain. The domain classifier $\mathnormal{G}_d : \mathbbm{R} ^ D \to [0,1]$ is defined as:
%\begin{equation}
%    \mathnormal{G}_d(\mathnormal{G}_f(x)) = \sigma (u^T \mathnormal{G}_f(x) + b)
%\end{equation}
%where $u$ is linear transformation and $b$ is bias term. The resulting loss function for our adaptive model is defined as:
%\begin{equation}
%\begin{aligned}
%\mathnormal{L}_d &(\mathnormal{G}_d(\mathnormal{G}_f(x_i)), d_i) =  d_i \log(\mathnormal{G}_d(\mathnormal{G}_f(x_i))^{-1}) \\
%& + (1 - d_i) \log((1- \mathnormal{G}_d(\mathnormal{G}_f(x_i)))^{-1}) \nonumber 
%\end{aligned}
%\end{equation}
%where $d_i$ is the domain label, indicating whether $x_i$ belongs to $\mathnormal{D}_s$ or $\mathnormal{D}_t$.

%% file: 04_experiments.tex
In this section, we first describe the datasets that have been used for the evaluation of our models, including our collected dataset. Then we present results of our proposed models on both (adaptive) causality identification and causality localisation.

\subsection{Datasets}
\begin{table}[t]
%\vspace{1ex}
    \centering
    \begin{tabular}{l c}
    \hline
        \#causality classes & 4\\
        average sentence length & 29.5 \\
        \#explicit causality & 9,092\\
        \#implicit causality & 616\\
        \#nested causality & 1,356\\
        \#non-causal & 3,936\\
        \hline
        \#Total sentences & 15,000 \\
        \hline
    \end{tabular}
    \caption{Statistics  on \dataset{MedCaus},  our collected dataset.}
    \label{tab:medcaus}
\end{table}
\paragraph{\dataset{MedCaus}} We introduce our medical causality dataset with 15,000 sentences. The process of collection and annotation of the sentences was followed by the guideline of \citet{hendrickx2009semeval}, including three main phases. In the first phase, sentences from medical articles of Wikipedia were randomly extracted. Using a wide variety of predefined causal connective words and patterns, we manually selected the sentences with potential causal relation and those without causal relation. In the second phase, the annotation instruction, multiple examples with different types of causal relation (i.e., explicit, implicit, nested, and non-causal) and different causal connective words were provided to the annotators. We asked four English-speaking graduate students to label the data accordingly. In the third phase, sentences with any disagreement that could not be resolved or were not clear in terms of causal relation were removed. To measure the level of agreement between our annotators, we give the same set of 1,000  sentences to the annotators. Using Fleiss Kappa measure \citep{fleiss1973equivalence} ($\mathcal{k}$), the level of agreement between our annotators has been  0.71, showing the reliability of the annotations. Table \ref{tab:medcaus} reports statistics about our collected dataset.

\paragraph{FinCausal} The dataset, which is extracted from financial news provided by QWAM~\footnote{http://www.qwamci.com/}, includes different sets for both tasks of causality identification and localisation. For the former, it includes 22,058 sentences, and for the latter task, 1,750 sentences were provided\footnote{http://wp.lancs.ac.uk/cfie/fincausal2020/}.

\paragraph{SemEval-10} We use SemEval-10 Task 8, which has 1,003 sentences with causal relation. From other relations of this dataset, we randomly select 997 sentences, totalling 2,000 sentences. The sentences from this dataset are selected from a wide variety of domains, however, unlike \dataset{MedCaus} the causal relations are indicated only between pair nominals. This dataset was used for both causality identification and causality localisation task~\citep{hendrickx2009semeval}.

\paragraph{BioCausal-Small} The dataset is a part of larger dataset~\footnote{The complete dataset is not publicly available.}, consisting of 2,000 biomedical sentences from which 1,113 have causal relations. The sentences from this dataset have been collected from biomedical articles of PubMed~\footnote{https://pubmed.ncbi.nlm.nih.gov}. Since this dataset only includes information about whether a sentence has causal relation~(regardless of the position of the cause and effect), it has been used for causality identification ~\citep{kyriakakis2019transfer}.

\subsection{Experimental Details}

For both GCE and ACE, we use Stanford CoreNLP~\citep{manning2014stanford} to generate the dependency parsing tree for each sentence. We use the pre-trained 300-dimensional Glove vectors~\cite{pennington2014glove} to initialise the embedding layer of our model. The hidden size for LSTM and the output feedforward layers is set to 100. We use the standard max pooling function for the pooling layer. Also, for all non-linearities in our model, we use Tanh function. A dropout ratio of $p = 0.5$ has been applied to all layers except for the last layer of GCN, for regularisation purposes.

For training of GCE, we split the data into train, development, and test set with the ratio of 60:20:20. For both models, we use batches of size 50. We train the model for 100 epochs, using Adamax optimiser. We use a decay rate of $0.9$ if the $F1$ score of development set does not increase after each epoch. The reported results are micro-averaged precision, recall, and F1 score. All the hyperparameter and training settings were kept the same as reported above for other models for comparison. The original GCN and C-GCN model~\citep{zhang2018graph}, which have been used as baselines for experiment, use the Named Entity Recognition and Part of Speech Tagging embeddings of the related named entity as input to the model. Since identifying causal relation is not limited to named entities only, to be able to adjust baseline models to our experiment setup, we trained these model without the aforementioned embeddings.  

\begin{table*}[!ht]
    \centering
    \begin{tabular}{l c c c c c c}
        &  \multicolumn{3}{c}{\dataset{MedCaus} } & \multicolumn{3}{c}{FinCausal} \\
        \hline 
        Model  & P & R & F1  & P & R & F1\\
        \hline
        P-Wiki~\citep{hidey2016identifying} & 74.4 & 74.4 & 74.4 & 54.0 & 54.0 & 54.0 \\
        bi-LSTM~\citep{martinez2017neural} & 84.2 & \textbf{97.8} & 90.5 & 81.3 & 77.0  & 79.1 \\
        GCN~\citep{zhang2018graph} & 90.8 & 94.6 & 92.7 & 85 & 74.8 & 79.6 \\
        C-GCN~\citep{zhang2018graph} & 91.2 & 94.9 & 93.0 & \textbf{86.1} & 68.7 & 76.4 \\
        \hline
        GCE & \textbf{92.5} & 94.0 & \textbf{93.2} & 84.8 & \textbf{83.3} & \textbf{84} \\
        \hline
    \end{tabular}
    \caption{Results of our proposed method on Task1, causality identification, compared with the baseline approaches on the \dataset{MedCaus} and FinCausal dataset.}
    \label{tab:Task1}
\end{table*}

\begin{table*}[!ht]
    \vspace{-3ex}
    \centering
    \begin{tabular}{l c c c c c c}
        &  \multicolumn{3}{c}{\dataset{MedCaus} } & \multicolumn{3}{c}{FinCausal} \\
        \hline
        Model  & P & R & F1  & P & R & F1 \\
        \hline
        bi-LSTM-CRF~\citep{martinez2017neural} & \textbf{77.4} & 69.9 & 73.4 & \textbf{82.4} & 65.0 & 72.7 \\
        GCN-CRF~\citep{zhang2018graph} & 31.9 & 46.8 & 37.9 & 66.1 & 55.5 & 60.3 \\
        C-GCN-CRF~\citep{zhang2018graph} & 72.5 & \textbf{75.9} & 74.1 & 76.3 & 68.8 & 72.3 \\
        S-LSTM-CRF~\citep{lample2016neural}  & 58.6 & 64.0 & 61.2 & 61.5 & 29.7 & 40.0 \\
        ELMO-CRF~\citep{peters2018deep} & 48.5 & 78.9 & 60.1 & 71.8 & 61.3 & 66.1 \\
        \hline
        GCE & 76.3 & 73.6 & \textbf{74.9} & 79.2 & \textbf{69.8} & \textbf{74.2} \\
        \hline
    \end{tabular}
    \caption{Results of our proposed method on Task2, causality localisation, compared with the baseline approaches on the \dataset{MedCaus} and FinCausal dataset.}
    \label{tab:Task2}
\end{table*}

\subsection{Task1: Causality identification}
In this section, we report the results on the task of identifying whether a sentence includes any causal relation or not. For this purpose, we use \dataset{MedCaus} and FinCausal to compare our GCE-based classifier (c.f. \S 3.1) with existing  methods for  causality identification. We divide the dataset into train/test/validation sets based on the ratio 60:20:20.

We compare our model to the P-Wiki \citep{hidey2016identifying}, which is a rule-based method, and bi-LSTM \citep{martinez2017neural}. Furthermore, since this task is closely related to the task of relation extraction, we compare our model to GCN, and C-GCN \citep{zhang2018graph}, which use dependency tree information of the sentence.

The results are reported in Table \ref{tab:Task1}. Our GCE-based classifier achieves the highest F1 and precision score on \dataset{MedCaus}, amongst all the models, followed closely by C-GCN. However, bi-LSTM shows the highest recall score. On FinCausal, our proposed model achieves the highest F1 and recall score, comparatively, while C-GCN hits the highest score on precision. Given the complexity and ambiguity of the projection of causal relation in natural language, taking both semantic and syntactic relations of a sentence improves the model. Hence, as suggested by the results, using both contextualised representation of a sentence and dependency relations of tokens of a sentence enriches the model, and results in obtaining more accurate prediction of causal relations.

\begin{table*}[!ht]
    \centering
    \small
    \begin{tabular*}{\textwidth}{l c c c c c c c c c}
        & \multicolumn{3}{c}{\dataset{MedCaus} $\to$ BioCausal} & \multicolumn{3}{c}{\dataset{MedCaus} $\to$ SemEval} & \multicolumn{3}{c}{\dataset{MedCaus} $\to$ FinCausal}\\ [3px]
        \hline
        Models & P & R & F1 & P & R & F1 & P & R & F1\\[3px]
        \hline
        bi-LSTM~\citep{martinez2017neural} & 76.1 & 57.6 & 66.0 & 82.5 & 62.8 & 71.3 & 47.6 & 8.3 & 14.2 \\ [3px]
        GCN~\citep{zhang2018graph} & 75.4 & 51.0 & 60.8 & 78.4 & 67.6 & 72.6 & 49.2 & 53.2 & 51.1 \\[3px]
        C-GCN~\citep{zhang2018graph} & 71.3 & 42.9 & 55.3 & 84.1 & 70.8 & 76.9 & 48.7 & 52.3 & 50.4 \\[3px]
        \hline
        bi-LSTM+DA & 75.6 & 58.9 & 66.2 & 81.6 & 69.5 & 75.1 & 47.9 & 61.1 & 53.7 \\[3px]
        GCN+DA & 72.8 & 70.3 & 71.5 & 82.9 & 66.7 & 73.9 & 46.8 & 57.4 & 51.6 \\[3px]
        C-GCN+DA & 78.4 & 55.5 & 65.1 & 81.9 & 71.0 & 76.1 & \textbf{49.1} & 54.6 & 52.7 \\[3px]
        \hline
        CDAN~\citep{long2018conditional} & \textbf{85.5} & 50.1 & 63.8 & \textbf{84.6} & 73.8 & 78.8 & 43.6 & 53.3 & 48.0 \\[3px]
        CDAN-E~\citep{long2018conditional} & \textbf{83.8} & 55.0 & 66.4 & 81.2 & 74.2 & 77.6 & 48.3 & 63.3 & 54.8 \\ [3px]
        \hline
        ACE & 74.3 & \textbf{77.1} & \textbf{76.7} & \textbf{84.4} & \textbf{74.2} & \textbf{79.0} & 47.4 & \textbf{74.0} & \textbf{57.8}\\[3px]
        \hline
    \end{tabular*}
    \caption{Results on our proposed adaptive causality encoder compared with the baselines for the task of causality identifier. The source dataset is \dataset{MedCaus}. BioCausal, SemEval, and FinCausal are considered as the target dataset.}
    \label{tab:da-task1}
\end{table*}

\begin{table*}[!ht]
    \centering
    \begin{tabular}{l c c c c c c}
        &  \multicolumn{3}{c}{\dataset{MedCaus} $\to$ SemEval} & \multicolumn{3}{c}{\dataset{MedCaus} $\to$ FinCausal}\\
        \hline
        Models & P & R & F1 & P & R & F1\\
        \hline
        bi-LSTM~\citep{martinez2017neural} & 16.3 & 52.2 & 24.9 & \textbf{64.1} & 16.1 & 25.8 \\
        GCN~\citep{zhang2018graph} & 8.8 & 29.6 & 13.5 & 41.6 & 40.9 & 41.0 \\
        C-GCN~\citep{zhang2018graph} & 18.9 & 47.6 & 27.1 & 63.8 & 13.0 & 21.6 \\
        \hline
        bi-LSTM+DA & \textbf{51.2} & 42.0 & 46.2 & 44.9 & 39.4 & 41.9 \\
        GCN+DA &  9.1 & 25.5 & 13.4 & 39.6 & 45.2 & 42.2 \\
        C-GCN+DA &  45.1 & 42.1 & 43.6 & 40.0 & \textbf{45.5} & 42.6 \\
        \hline
        CDAN~\citep{long2018conditional} &  40.9 & 49.7 & 44.8 & 36.9 & 42.8 & 39.6 \\
        CDAN-E~\citep{long2018conditional} &  47.3 & 40.6 & 43.7 & 36.8 & 42.6 & 39.5 \\
        \hline
        ACE & 42.3 & \textbf{53.6} & \textbf{47.3} & 42.2 & 43.2 & \textbf{42.7} \\
        \hline
    \end{tabular}
    \caption{Results on our proposed adaptive causality encoder compared with the baselines for the task of causality localisation. The source dataset is \dataset{MedCaus} and target dataset are SemEval and FinCausal.}
    \label{tab:da-task2}
\end{table*}

\subsection{Task2: Causality Localisation}

This section covers the results of the performance of our proposed model, compared to other models, in terms of extracting cause and effect from textual data. \dataset{MedCaus} and FinCausal are used in this task for evaluation purposes. Each dataset are split into train/test/validation with the ratio of 60:20:20. For comparison, we report the results of the performance of each model in labelling each token with the proper tag. For this purpose, precision, recall, and F1 score are reported.

Similar to the Task1, we compare our model to bi-LSTM \cite{martinez2017neural}, GCN, and C-GCN \citep{zhang2018graph}. Also, we compare our model to the proposed model of~\citet{lample2016neural}, with two variations of using S-LSTM and ELMO~\citep{peters2018deep} for contextual embedding.

The results of causality localisation are reported in Table \ref{tab:Task2}. The experiments on \dataset{MedCaus} show that while bi-LSTM-CRF achieves better results in precision, it fails to gain high recall. On the hand C-GCN-CRF achieves highest recall, followed closely by our model. However, in F1 score, our model, outperforms the baselines. On FinCausal, bi-LSTM-CRF achieves the highest precision. However, our model achieves better recall and F1 score.

\subsection{Results of ACE}

%In this section, we present the results of our ACE model on the task of adaptive causality identification. To this end, we consider \dataset{MedCaus} as the source domain, and \emph{SemEval-10} and \emph{BioCausal} as two target sources. We compare our model to bi-LSTM \citep{martinez2017neural}, bi-LSTM + domain classifier, GCN, GCN + domain classifier, C-GCN, C-GCN + domain classifier \citep{zhang2018graph}, and a stat-of-the-art approach of conditional adversarial domain adaptation (CDAN-E) \citep{long2018conditional}.

In this section, we present the results of our ACE model on the task of adaptive causality identification and causality localisation. To this end, we consider \dataset{MedCaus} as the source domain, and SemEval-10, BioCausal, and \dataset{FinCausal} as target domains~\footnote{Since BioCausal does not provide tags of cause and effect, this dataset was not used for domain adaptive causality localisation.}. We compare our model to bi-LSTM~\citep{martinez2017neural}, GCN and C-GCN~\citep{zhang2018graph} as the baselines, their domain adaptive versions (indicated with ``+DA'' in the tables), and a state-of-the-art approach of conditional adversarial domain adaptation (CDAN and CDAN-E) \citep{long2018conditional}.

\begin{figure*}[!ht]
  \subfloat[\dataset{MedCaus} $\to$ SemEval]{
	\begin{minipage}[c][1\width]{
	   0.16\textwidth}
	   \centering
	   \includegraphics[width=1\textwidth]{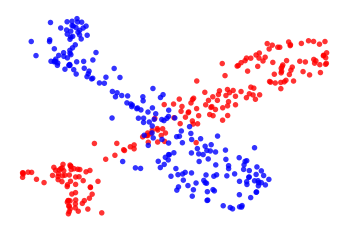}
	\end{minipage}
	\begin{minipage}[c][1\width]{
	   0.16\textwidth}
	   \centering
	   \includegraphics[width=1\textwidth]{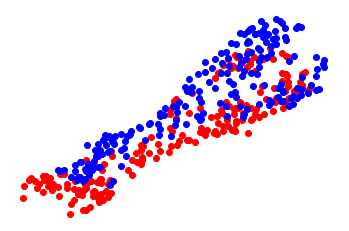}
	\end{minipage}}
 \hfill
 \subfloat[\dataset{MedCaus} $\to$ BioCausal]{
	\begin{minipage}[c][1\width]{
	   0.16\textwidth}
	   \centering
	   \includegraphics[width=1\textwidth]{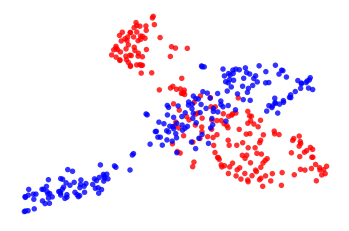}
	\end{minipage}
	\begin{minipage}[c][1\width]{
	   0.16\textwidth}
	   \centering
	   \includegraphics[width=1\textwidth]{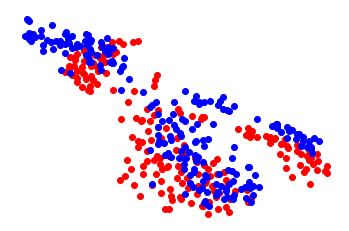}
	\end{minipage}}
 \hfill 
 \subfloat[\dataset{MedCaus} $\to$ FinCausal]{
	\begin{minipage}[c][1\width]{
	   0.16\textwidth}
	   \centering
	   \includegraphics[width=1\textwidth]{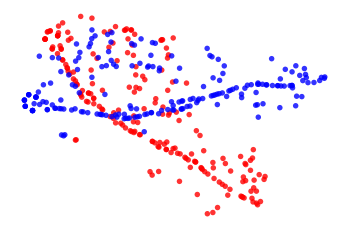}
	\end{minipage}
	\begin{minipage}[c][1\width]{
	   0.16\textwidth}
	   \centering
	   \includegraphics[width=1\textwidth]{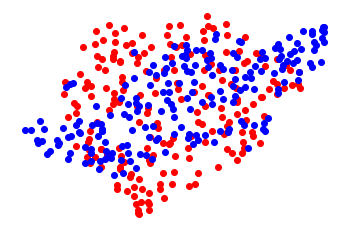}
	\end{minipage}}
\vspace{10pt}
\caption{t-SNE visualisation of the domain adaptation task~(with perplexity of 20 and Principal Decomposition Analysis~(PCA)~\cite{wold1987principal} initialisation). The source and target data are shown in red and blue, respectively. In each section, features before and after applying ACE are represented,on left and right side, respectively. }
\label{fig:tsne}
\end{figure*}

Table \ref{tab:da-task1} summarises the results of our experiments of domain adaptive causality identification. As it can be seen, adding the domain adaptive strategy to the baselines improves their performance on all target datasets, by up to 39\%. Furthermore, while CDAN achieves a better precision, it fails to balance the recall and performs poorly in terms of F1 score. On the other hand, our model (ACE; c.f. \S 3.2), outperforms all of the other models in recall and F1 score. 

The results on applying domain adaptation method for the task of causality localisation is reported in Table \ref{tab:da-task2}. Applying our proposed domain adaptive model has improved the recall and F1 score of the baselines on both target datasets. While other models achieve better precision scores, our model consistently gains a better recall and F1 score, showing the superiority of our approach.

\begin{table*}[!ht]
    \centering
    \small
    \begin{tabular}{ || c | c | p{12cm} c||}
    \hline\hline
         \multirow{14}{*}{\rotatebox[origin=c]{90}{Causality Identification}} & \multirow{3}{*}{\dataset{MedCaus}} & 1. Severe narrowings may cause chest pain~(angina) or breathlessness during exercise or even at rest. & \checkmark \\ [2px]
          & & 2. When the floor of the mouth is compressed, air is forced into the lungs. & $\times$
          \tabularnewline\cline{2-4}
          & \multirow{4}{*}{SemEval} & 1. Mechanical faults caused delays and cancellations on Wellington's suburban train services this morning & $\times$ \\ [2px]
          & & 2. The overall damage caused by the destruction of land and property for the Wall's construction has taken many years to recover further. & \checkmark 
          \tabularnewline\cline{2-4}
          & \multirow{4}{*}{FinCausal} & 1.  Thomas Cook, one of many world's largest journey corporations, was based in 1841 to function temperance day journeys, and now has annual gross sales of 39 billion. & $\times$ \\[2px]
          & & 2. The judge's decision converted the arbitration award to a legal judgement and the sum, including interest accrued since 2013, soared to more than \$9 billion. & \checkmark 
          \tabularnewline\cline{2-4}
          & \multirow{3}{*}{BioCausal} & 1. For cost and convenience reasons other altered fractionation schedules have been adopted in routine practice. & $\times$ \\[2px]
          & & 2. The sequential technique also minimises the incidence of iris bleeding. & \checkmark \\
          \hline\hline
          \multirow{10}{*}{\rotatebox[origin=c]{90}{Causality Localisation}} & \multirow{4}{*}{\dataset{MedCaus}} & 1. \textcolor{Maroon}{\textbf{\uline{A high rate of consumption}}} can also lead to \textcolor{BlueViolet}{\textbf{\dashuline{cirrhosis, gastritis, gout, pancreatitis, hypertension, various forms of cancer, and numerous other illnesses.}}} & \\
          & & 2. \textcolor{Maroon}{\textbf{\uline{The phlegm produced by catarrh}}} may either discharge or cause \textcolor{BlueViolet}{\textbf{\dashuline{a blockage}}} that may become chronic. & 
          \tabularnewline\cline{2-4}
          & \multirow{3}{*}{SemEval} & 1. He took a shower after using hair cream to avoid \textcolor{BlueViolet}{\textbf{\dashuline{skin irritation}}} from the \textcolor{Maroon}{\textbf{\uline{chemicals in the product}}}. & \\[2px]
          & & 2. A \textcolor{Maroon}{\textbf{\uline{cigarette}}} set off a \textcolor{BlueViolet}{\textbf{\dashuline{smoke alarm}}}. & 
          \tabularnewline\cline{2-4}
          & \multirow{3}{*}{FinCausal} & 1. \textcolor{BlueViolet}{\textbf{\dashuline{The DGR in the Roth is lower at 5.4\%}}} due primarily to its \textcolor{Maroon}{\textbf{\uline{holding of REITs}}}. & \\[2px]
          & & 2.  \textcolor{BlueViolet}{\textbf{\dashuline{Company tax receipts}}} were \$4.6 billion higher than predicted, mainly due to \textcolor{Maroon}{\textbf{\uline{mining profits}}}, but Mr Frydenberg could not say how much was due to strong iron ore demand. & \\[2px]
          \hline \hline
    \end{tabular}
    \caption{Examples of the performance of our proposed model in causality identification and localisation. The top section of the table provides examples for the first task. The \checkmark and $\times$ indicate identification of causal and no causal relation in the sentence, respectively. The bottom part of the table presents examples for the second task. We have used underline with red colour and dashed underline with blue colour to show cause and effect respectively.}
    \label{tab:example}
\end{table*}

\textbf{Visualisation} Figure \ref{fig:tsne} visualises the effect of applying our proposed domain adaptation module (ACE; c.f. \S 3.2), to different target datasets. The extracted features ($f_{\theta_{enc}}$) of the source and target datasets are visualised using t-distributed Stochastic Neighbour Embedding (t-SNE)~\citep{maaten2008visualizing}. The source and target datasets are shown in red and blue, respectively. In each sub-figure, the features before and after applying ACE are represented on the left and right side, respectively. It is clear that, where the source and target domains data have different distributions, ACE matches the distributions, which greatly helps with improving the performance on the target data.

\subsection{Qualitative Analysis}

In this section, we demonstrate the capability of our proposed models in addressing the tasks of causality identification and localisation. To this end, for each task, two sentences from each dataset are presented in Table \ref{tab:example}. The top section of the table provides examples for causality identification. The bottom section presents example for causality localisation. The examples suggests that our proposed models perform accurately on datasets with different distributional features.

%% file: 05_conclusion.tex
In this work, we propose a new dataset for the task of causal identification and causal extraction from natural language text. 
We further propose a neural-based model for textual causality identification and localisation, which makes use of dependency trees.
We then make use of 
adversarial training
to adapt the causality identification and localisation models to new domains. 
Empirical results show that our method outperforms state-of-the-art models and their adapted versions.